\pdfoutput=1

\documentclass[11pt]{article}

\usepackage{naacl2021}

\usepackage{times}
\usepackage{latexsym}

\usepackage[disable]{todonotes}
\usepackage{graphicx}
\usepackage{booktabs}
\usepackage{subcaption}
\usepackage{adjustbox}
\usepackage{mathtools}
\usepackage{array}
\usepackage{amsmath}
\usepackage{amsfonts}
\usepackage{multirow}
\usepackage{tabularx}
\usepackage{hyperref}

\usepackage[T1]{fontenc}

\usepackage[utf8]{inputenc}
\usepackage{csquotes}

\usepackage{microtype}

\author{Kareem Ahmed, Eric Wang, Guy Van den Broeck \and Kai-Wei Chang\\
Computer Science Department\\
University of California, Los Angeles\\ 
Los Angeles, CA, USA\\
\texttt{\small ahmedk@cs.ucla.edu,ericzxwang@ucla.edu,\{guyvdb, kwchang\}@cs.ucla.edu} 
}

\usepackage{siunitx,etoolbox}
\newrobustcmd{\B}{\bfseries}
\title{Leveraging Unlabeled Data for Entity-Relation Extraction through Probabilistic Constraint Satisfaction}

\begin{document}
\maketitle
\begin{abstract}

We study the problem of entity-relation extraction in the presence of symbolic domain knowledge. Such knowledge takes the form of an ontology defining relations and their permissible arguments. Previous approaches set out to integrate such knowledge in their learning approaches either through self-training, or through approximations that lose the precise meaning of the logical expressions. By contrast, our approach employs semantic loss which captures the precise meaning of a logical sentence through maintaining a probability distribution over all possible states, and guiding the model to solutions which minimize any constraint violations. With a focus on low-data regimes, we show that semantic loss outperforms the baselines by a wide margin.

\end{abstract}

\section{Introduction}
With an abundance of textual data being produced, relation extraction --- the task of identifying relations between named entities --- offers a means by which structured knowledge can be extracted from unstructured text and used to populate relational databases. However, such relation extraction models typically require large amounts of labeled data, which can be hard and expensive to acquire. Therefore, increasing attention has been going into semi-supervised learning with the hope that, given enough knowledge about the structure of the problem, such approaches can be competitive with their fully-supervised counterparts.

In this work, we take advantage of the domain knowledge present when dealing with relational data, and define a set of relations and their permissible argument types (e.g., the relation \textit{kill}, in its literal sense, does not admit inanimate entity types as either of its arguments). We encode such an ontology as Boolean logic sentences, the conjunction of which we refer to as our constraint.

Using semantic loss \cite{xu18}, our approach maintains a probability distribution over all states predicted by the model, and penalizes the model for any probabilities it assigns to states violating the constraint. Calculating the probabilities of each state is exact --- capturing the precise \textit{meaning} of the logic expression --- and efficient, only adding a linear (in the size of the compiled circuit; which is small in practice) overhead to our training.

While not the first to exploit unlabeled data under general constraints for entity-relation extraction, our approach enjoys several advantages to other commonly used approaches. The Constraint-Driven Learning (CoDL) framework \cite{chang2007} necessitates solving an integer linear program (ILP) for every sample in the unlabeled set, for every epoch, with the goal of providing the model with pseudo-labeled examples that respect the constraint, a computationally expensive process. CoDL collapses the probability distribution maintained by the model over all states to a single point: the most probable state, often reinforcing the model's incorrect predictions.

Another class of approaches attempts to reduce logical constraints into differentiable, arithmetic objectives by substituting logical operators with their fuzzy t-norms and logical implications with simple inequalities. A downside of this fuzzy relaxation is that the logical sentences lose their precise meaning as the learning objective becomes a function of the syntax of the logical expression rather than its semantics. 

Our experiments show how our approach of injecting logical constraints into the learning process offers significant improvements over the baselines, leading to more logically consistent predictions. We echo previous findings that the latter could be augmented with constraints at inference using ILP to further improve the performance. However, we find that transductive training is competitive with ILP, and could therefore be used as a proxy for inference-time constraints, simplifying the training pipeline. We make our code available online.

\section{Related Work}
Our work builds upon a body of literature on semi-supervised approaches that use structural constraints as a source of indirect supervision, and have been applied to several NLP tasks. \citet{chang2007} introduce the Constraint-Driven Learning (CoDL) framework which improves the model by generating feedback through labeling unlabeled examples subject to a hard constraint. \citet{Ganchev10} develop a CoDL-like approach, with the relaxation that constraints be satisfied in expectation, and finds the best variational distribution satisfying these constraints. Along the same lines, the generalized expectation framework \cite{Mann07, Mann08, Mann10} specifies a set of linear constraints and augments the log-likelihood objective with a penalty term: the expected deviation of the model's output from the constraint specified.

More in line with our work, various deep learning techniques have been proposed to enforce either arithmetic constraints \cite{pathak15, Marquez-Neila17} or logical constraints \cite{rocktaschel15, hu16, demeester16, stewart16, minervini17, diligenti17, Donadello17, fischer19a} on the output of a neural network, through reducing logical constraints into differentiable, arithmetic objectives by replacing logical operators with their fuzzy t-norms and logical implications with simple inequalities. While semantic loss is exact, and captures the precise meaning of a logical sentence regardless of its syntax, such fuzzy relaxations tend to lose the precise meaning of the logical sentences and consequently, the learning objective becomes a function of the syntax rather than the semantics.

\section{Methodology}
\begin{figure}
    \centering
    \begin{subfigure}[]{\columnwidth}
\tikzset{every picture/.style={line width=0.75pt}} 

\begin{tikzpicture}[x=0.75pt,y=0.75pt,yscale=-1,xscale=1]

\draw   (175,90) -- (468.5,90) -- (468.5,145) -- (175,145) -- cycle ;
\draw (321,119) node  [align=left] {$\displaystyle  \begin{array}{{>{\displaystyle}l}}
\alpha \ \coloneqq ( Kill\ \Longrightarrow Person_{s} \ \land Person_{o})\\
\ \ \ \ \ \land\ \ ( LivesIn\Longrightarrow Person_{s} \ \land Location_{o})\\
\end{array}$};
\end{tikzpicture}
    \end{subfigure}%
    \caption{An example constraint $\alpha$ in Boolean logic: a relation \textit{Kill} can only have a \textit{Person} for both its subject and its object. Similarly, a relation \textit{LivesIn} can only gave a \textit{Person} as its subject and a \textit{Location} as its object.}
    \label{fig:example_constraint}
\end{figure}
Consider the following instance of the relation extraction problem, with two possible relations: \textit{Kill} and \textit{LivesIn}, and two possible entity types: \textit{Person} and \textit{Location}. Consider the following sentence:
\begin{displayquote}
\centering
Oswald shot JFK
\end{displayquote}
\noindent and assume the following model assignments: Oswald is predicted as a \textit{Person} with probability $0.3$ and a \textit{Location} with probability $0.7$; JFK is predicted as a \textit{Person} with probability $0.1$ and a \textit{Location} with probability $0.9$; the relation is predicted as \textit{Kill} with probability $0.6$, and as \textit{LivesIn} with probability $0.4$.

Denote by a \textit{state} any (subject, relation, object) triple. We define the \textit{probability of a state} to be the product of probabilities of the classes in that state i.e. the product of probabilities of the subject, relation and object. More formally,
\begin{equation*}
    p((e_i, r_m, e_j)) = p_\phi(e_i) \times p_\theta(r_m) \times p_\phi(e_j)
\end{equation*}
for a subject $i$, a relation $m$, and an object $j$, where $\phi$ and $\theta$ denote the parameters of the entity recognition and relation extraction modules, respectively. This induces a probability distribution over all possible states, with the sum of probabilities of all possible states adding to $1$. Continuing with our example, given the model predictions above, we compute the following distribution, where each row denotes a state and its corresponding probability:

\begin{table}[!htbp]
\begin{center}
\resizebox{\columnwidth}{!}{%
\begin{tabular}{lc||lc}
\toprule
state & prob & state & prob \\
\midrule
(Per, Kill, Per) &0.02 &(Loc, Kill, Loc) &0.38\\
(Per, LivesIn, Loc) &0.11 & (Per, LivesIn, Per) &0.01\\
(Per, Kill, Loc) &0.16 & (Loc, LivesIn, Per) &0.03\\
(Loc, Kill, Per) &0.04 & (Loc, LivesIn, Loc) &0.26\\

\bottomrule
\end{tabular}
}
\end{center}
\vskip -0.1in
\end{table}

Now consider the constraint in Figure \ref{fig:example_constraint}. As per the constraint, the relation \textit{Kill} can only co-occur with a subject \textit{Person} and an object \textit{Person}. Similarly, the relation \textit{LivesIn} can only co-occur with a subject \textit{Person} and an object \textit{Location}. That is, subject to our constraint, the only \textit{valid} states  are \textit{(Per, Kill, Per)} and \textit{(Per, LivesIn, Loc)}. Following \citet{xu18}, we seek to maximize the probability the model assigns to such \textit{valid} states, or equivalently, minimize the probability of predicting an \textit{invalid} state. Our loss is then defined in terms of the probability that the constraint is \textit{satisfied}: a summation over the probabilities of all valid states. Taking the negative logarithm, we get a loss term proportional to that of the cross-entropy loss, and  attains a value of $0$ when the constraint is always satisfied. Formally, the semantic loss function is
\begin{equation}
    L^s(\alpha, \mathbf{p}) = - \log \sum_{(e_i, r_m, e_j) \; \textit{sat} \; \alpha} \mathbf{p}((e_i, r_m, e_j)),
\end{equation}
where $\alpha$ denotes our constraint, $\mathbf{p}$ denotes a vector of probabilities over states, and \textit{$(e_i, r_m, e_j) \; \text{sat} \; \alpha$} denotes that a state with the subject $i$, relation $m$ and object $j$ is valid under the constraint $\alpha$. Consequently, our example would yield a loss of $- \log(0.02 + 0.11) = - \log(0.13) = 2.04 $.

\label{sec:model}
\subsection{Architecture}
Similar to prior works in the literature on entity-relation extraction \cite{miwa2014, miwa2016, gupta2016, Li16, Li17, zhang2017, Adel17, bekoulis2018a, bekoulis2018b, nguyen2019, Li19}, we adopt a end-to-end approach to recognizing the named entities in a sentence along with their pairwise relations. Contextual embeddings are first produced for every token in the sentence. These are then fed to a named entity recognition (NER) module, which outputs a vector of probabilities denoting the per-class probability. The output of the NER module is then passed as an input, along with the contextual embeddings, as well as subject and object index embeddings, to a relation classifier, which outputs the corresponding relation. We detail the architecture of each of the aforementioned modules below.
\paragraph{Contextualized Encoder} Our model encodes words $w_i$ of an input sequence $s = (w_1, w_2, \ldots, w_n)$ using the pretrained BERT$_{Base}$ model \cite{devlin19} to produce a sequence of embedding vectors, $\mathbf{e_1, e_2, \ldots, e_n}$, where $\mathbf{e_i}$ denotes the contextualized word embedding of token $w_i$.
\paragraph{Named Entity Recognition and Relation Extraction Modules} As we assume we only have a small labeled set, we fix the contextualized encoder, and use BERT + LSTM as proposed in \citet{Wadden19},  where the LSTM parameters are trained together with task specific layers. Both the Named Entity Recognition (NER) and Relation Extraction (RE) modules are provided with a list of contextualized embeddings as their input, with the RE module additionally provided the NER predictions. We jointly minimize the NER cross-entropy loss, RE cross-entropy loss and the semantic loss.

\begin{table*}[!htb]
\centering
\resizebox{\textwidth}{!}{%
\begin{tabular}{lccc|ccc|ccc|ccc|ccc|ccc|ccc}
\toprule
\# Labels      &\multicolumn{3}{c}{3}  &\multicolumn{3}{c}{5}  &\multicolumn{3}{c}{10} &\multicolumn{3}{c}{15} &\multicolumn{3}{c}{25} &\multicolumn{3}{c}{50} &\multicolumn{3}{c}{75}\\
{Inf}     &N &I &T                &N &I &T                &N &I &T                &N &I &T                &N &I &T                &N &I &T                &N &I &T\\
\midrule
base         &35.2 &49.3 &35.2             &38.3 &54.3 &38.3             &45.0 &58.8 &45.0             &46.6 &60.6 &46.6             &55.0 &65.5 &55.0             &60.3 &70.1 &60.3             &63.5 &73.0 &63.5\\
CoDL        &38.6 &52.2 &36.6             &44.1 &56.3 &45.3             &48.5 &60.3 &49.0             &50.0 &61.9 &49.6             &58.3 &\B68.4\rlap{$^*$} &55.7           &62.9 &70.6 &63.1             &66.5 &73.9 &65.3\\   
t-norm      &39.0 &50.7 &40.4             &45.1 &55.8 &42.3             &50.3 &58.1 &51.4             &53.3 &62.1 &56.4             &60.4 &67.2 &59.2             &63.9 &70.4 &65.8             &66.5 &73.3 &67.0\\   
sl (ours)   &\B44.3 &\B55.2\rlap{$^*$} &\B45.4       &\B52.4 &\B60.9\rlap{$^*$} &\B50.5       &\B55.2 &\B62.9\rlap{$^*$} &\B56.4       &\B58.9 &\B66.0\rlap{$^*$} &\B60.1       &\B60.0 &68.0 &\B62.5         &\B65.4 &\B73.2\rlap{$^*$} &\B68.1       &\B67.2 &\B74.2\rlap{$^*$} &\B68.2\\
\midrule
\midrule
base           &4.9 &7.6 & 4.9           &7.2 &11.9&7.2              &13.7 &18.9 & 13.7          &15.1 &19.0 & 15.1          &21.6 &26.5 & 21.6          &29.0 &33.6 & 29.0         &33.0 &37.4 & 33.0 \\
CoDL           &7.7 &8.9 &4.8            &12.8 &14.2 &12.3           &16.2 &17.7 &17.3           &17.6 &18.4 &17.7           &27.0 &28.3 &22.4           &32.9 &34.6 &32.4          &37.2 &38.5 &35.3\\
t-norm         &8.9 &9.5 &9.5          &14.5 &15.5 &10.2           &19.2 &14.2 &19.6           &21.8 &23.1 &24.9           &\B30.2 &30.7 &29.4           &34.1 &35.6 &37.3          &37.4 &38.6 &39.5\\
sl (ours)      &\B12.9 &\B14.0 &\B14.1\rlap{$^*$}   &\B20.3 &\B22.3\rlap{$^*$} &\B18.5     &\B24.6 &\B25.6 &\B26.2\rlap{$^*$}     &\B30.1 &\B31.3 &\B31.6\rlap{$^*$}     &29.2 &\B31.9 &\B33.6\rlap{$^*$}     &\B36.5 &\B39.4 &\B40.5\rlap{$^*$}    &\B38.8 &\B42.0\rlap{$^*$} &\B40.3\\
\bottomrule
\end{tabular}%
}
\vskip -0.1in
\caption{Experimental results for joint entity-relation extraction on ACE05. The upper and lower sub-tables correspond to \textit{avg-f1} and \textit{tri-f1}, respectively. \textbf{Inf} indicates no constraints during inference (N), use of ILP during inference (I) or the transductive setting (T). We averages across 3 different runs. Best results in a column and block are boldface and asterisked, respectively. We point out that better results in one metric do not translate to better results in the other, and focus on \textit{tri-f1} in our analysis. Full results, including std.~error, available in the appendix.}
\label{table:results}
\end{table*}

\section{Experiments}

\paragraph{Dataset and Evaluation Criteria}
We conduct experiments using the Automatic Content Extraction (ACE) 2005 corpus \cite{walker2006ace} that defines an ontology over 7 entities and 18 relations. For each method, we report two evaluation metrics: (1) the average of the F1-scores of the relations and their arguments, which we denote \textit{avg-f1}, and (2) the F1-score of relation triples, which we denote \textit{tri-f1}, where a triple is considered correct only if both the relation and its arguments have been correctly classified.

We compare our proposed approach against three different baselines:

\paragraph{base} follows the model detailed in Section \ref{sec:model}, but does not make use of unlabeled data (i.e., the weight of semantic loss is set to $0$).

\paragraph{CoDL} denotes the Constraint Driven Learning framework by \citet{chang2007}. Given a small labeled set and a large unlabeled set, the base model is trained on the labeled set until convergence and then repeatedly: (1) Uses constraints and the learned model to label the instances in the unlabeled set. (2) Updates the model using the newly labeled data. We attempted to introduce constraints at different training stages, and achieved the best results by finetuning the best-performing model checkpoint.
min(1, rv / lv)
\paragraph{Product T-norm} follows the approach by \citet{rocktaschel15} which, similar to semantic loss, aims to maximize the probability that the model predictions satisfy a logical constraint. More specifically, the probability of any logical formula is computed recursively as:
\begin{align*}
    [\lnot \mathcal{A}] &= 1 - [\mathcal{A}]\\
    [\mathcal{A} \land \mathcal{B}] &= [\mathcal{A}][\mathcal{B}]\\
    [\mathcal{A} \lor \mathcal{B}] &= [\mathcal{A}] + [\mathcal{B}] - [\mathcal{A}][\mathcal{B}]\\
    [\mathcal{A} \implies \mathcal{B}] &= [\mathcal{A}]([\mathcal{B}] - 1) + 1
\end{align*}
where the $[\;\;]$ operator denotes the probability of the logical formula under the learned model. The above computation is exact when the arguments of the logical expression do not overlap. Otherwise, it can be thought of as an approximation, and becomes a function of the syntax, as well as the actual meaning of the logical formula. For example, the following constraints admit different probabilities:
\begin{align*}
    [\lnot Kill \lor Per \land Per]&=0.42,\\
    [Kill \implies Per \land Per]&=0.61.
\end{align*}
This is a problem, especially in case of larger constraints, from which semantic loss does not suffer.
\begin{figure}
    \includegraphics[width=\linewidth]{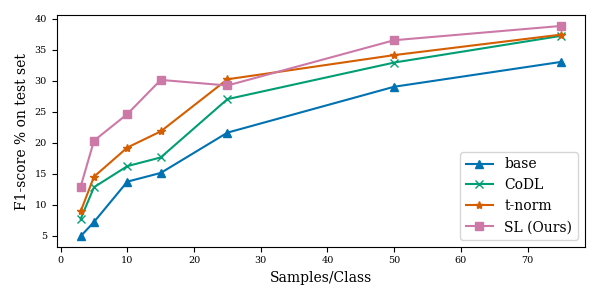}
    \caption{Comparing the performance (\textit{tri-f1}) of the different methods of injecting constraints to the baseline.}
    \label{fig:base_codl_sl}
\end{figure}
\section{Results and Discussion}
%
\textbf{Q1. Is injecting constraints into the learning process at all beneficial?} Our results show that injecting logical constraints into the learning process is \textit{critical}. This effect is especially pronounced in settings with very scarce labeled data. For instance, in settings with only 3 labeled samples per class, CoDL, product t-norm and semantic loss are seen to improve upon the baseline by $2.8\%$, $5.3\%$ and $8.0\%$, respectively, as can be seen in Table \ref{table:results}.\\

\noindent\textbf{Q2. Should we prefer one means of injecting logical constraints over another?} While all means of injecting logical constraints do improve upon a baseline that is oblivious to these logical constraints, not all of them perform equally well. Figure \ref{fig:base_codl_sl} shows that using semantic loss consistently outperforms CoDL and product t-norm across almost all data settings. We attribute such performance improvement to semantic loss (1) being exact, thereby avoiding the pitfalls of approximations which depend heavily upon the syntax of the logical expressions, and (2) maintaining a full posterior distribution over possible states throughout training, thereby incorporating the model's uncertainty about its predictions, rather than committing to the state most likely under the constraint.\\

\noindent\textbf{Q3. Can constraints at inference time replace constraints at training time?} To begin with, we focus on two sets of experiments, base-I and CoDL-N. The results in Table \ref{table:results} echo the conclusions in \citet{chang2007}: testing with constraints is more important than learning with constraints, especially as more labeled data is made available. That conclusion changes, however, once we compare base-I and SL-N: \textit{using semantic loss to inject constraints into the learning process leads to better performing models compared to base-I across all data settings.}\\

\noindent \textbf{Q4. Can we forgo constraints at inference time?} Not quite. As we can see in Table \ref{table:results}, \textit{learning and inference with constraints is superior to only learning with constraints}. We do also see, however, that \textit{transductive learning is competitive with enforcing constraints at inference time}. Therefore, it is possible to avoid using ILP and use transduction instead. This alleviates the need for implementing different techniques, and more importantly, circumvents the need to solve an expensive ILP at inference time.

\section{Conclusion}
Our work is a continuation of a long line of work on compensating for the lack of supervision in relation extraction using symbolic domain knowledge. We approach the problem in a principled manner: steering the model away from its incorrect beliefs through exactly, and efficiently tracking a distribution over the model's beliefs, and penalizing incorrect ones. We have shown such a principled approach to outperform competing methods in an experimental setup where labeled data range from extremely scarce to moderately scarce.

\section{Broader Impact}
Machine learning has become ubiquitous in our lives, taking part in everything from the mundane to the critical. These models, while achieving record shattering performances, quite often produce logically inconsistent outputs. By devising algorithms capable of injecting domain knowledge into a model's learned hypotheses, we are able to eliminate many of these logical inconsistencies, resulting in a more robust model. We note that, although our experiments show more consistent predictions, our model is in no way perfect. It is a result of a stochastically trained neural network, and may suffer from many of the shortcomings of such models. That is to say, a user must be aware of the limitations of such a system, especially when used as a proxy for people in decision making.

\bibliography{custom}
\bibliographystyle{acl_natbib}

\newpage
\appendix
\onecolumn
\pagebreak

\section{ACE05 Full Results}
\begin{table*}[!htb]
\centering
\resizebox{\columnwidth}{!}{%
\begin{tabular}{lccc|ccc|ccc|ccc|ccc|ccc|ccc}
\toprule
\# labels      &\multicolumn{3}{c}{3}  &\multicolumn{3}{c}{5}  &\multicolumn{3}{c}{10} &\multicolumn{3}{c}{15} &\multicolumn{3}{c}{25} &\multicolumn{3}{c}{50} &\multicolumn{3}{c}{75}\\
{Inf}     &N &I &T                &N &I &T                &N &I &T                &N &I &T                &N &I &T                &N &I &T                &N &I &T\\
\midrule
base        &35.2$_{0.7}$ &49.3$_{1.3}$ &35.2$_{0.7}$  &38.3$_{2.4}$  &54.3$_{1.7}$ &38.3$_{2.4}$  &45.0$_{1.0}$ &58.8$_{1.0}$ &45.0$_{1.0}$  &46.6$_{0.8}$ &60.6$_{0.4}$ &46.6$_{0.8}$  &55.0$_{2.1}$ &65.5$_{2.0}$ &55.0$_{2.1}$  &60.3$_{0.8}$ &70.1$_{1.1}$ &60.3$_{0.8}$  &63.5$_{0.6}$ &73.0$_{0.3}$ &63.5$_{0.6}$\\
CoDL        &38.6$_{3.4}$ &52.2$_{3.2}$ &36.6$_{0.6}$  &44.1$_{3.3}$ &56.3$_{3.1}$ &45.3$_{6.4}$   &48.5$_{3.0}$ &60.3$_{1.1}$ &49.0$_{1.6}$  &50.0$_{1.3}$ &61.9$_{0.3}$ &49.6$_{2.7}$  &58.3$_{2.6}$ &\B68.4\rlap{$^*$}$_{1.0}$ &55.7$_{1.7}$  &62.9$_{1.8}$ &70.6$_{2.1}$ &63.1$_{1.9}$  &66.5$_{0.9}$ &73.9$_{1.7}$ &65.3$_{1.9}$\\   
t-norm      &39.0$_{5.8}$ &50.7$_{4.8}$ &40.4$_{4.3}$  &45.1$_{1.6}$ &55.8$_{0.6}$  &42.3$_{0.8}$   &50.3$_{5.0}$ &58.1$_{4.9}$ &51.4$_{2.1}$  &53.3$_{6.7}$ &62.1$_{4.9}$ &56.4$_{1.1}$  &60.4$_{0.5}$ &67.2$_{0.5}$ &59.2$_{2.1}$  &63.9$_{2.3}$ &70.4$_{2.1}$ &65.8$_{1.2}$  &66.5$_{2.1}$ &73.3$_{1.7}$ &67.0$_{1.4}$\\   
sl (ours)   &\B44.3$_{5.7}$ &\B55.2\rlap{$^*$}$_{3.0}$&\B45.4$_{2.9}$   &\B52.4$_{2.4}$ &\B60.9\rlap{$^*$}$_{3.0}$ &\B50.5$_{0.2}$   &\B55.2$_{3.4}$ &\B62.9\rlap{$^*$}$_{2.7}$ &\B56.4$_{2.5}$  &\B58.9$_{1.2}$ &\B66.0\rlap{$^*$}$_{1.7}$ &\B60.1$_{1.0}$  &\B60.0$_{1.2}$ &68.0$_{1.0}$ &\B62.5$_{2.2}$  &\B65.4$_{0.5}$ &\B73.2\rlap{$^*$}$_{0.5}$ &\B68.1$_{0.9}$  &\B67.2$_{0.9}$ &\B74.2\rlap{$^*$}$_{0.8}$ &\B68.2$_{0.3}$\\
\midrule
\midrule
base        &4.9$_{1.1}$ &7.6$_{1.1}$ & 4.9$_{1.1}$   &7.2$_{1.8}$ &11.9$_{1.3}$ &7.2$_{1.8}$    &13.7$_{0.2}$ &18.9$_{1.7}$ & 13.7$_{0.2}$  &15.1$_{1.8}$ &19.0$_{1.5}$ & 15.1$_{1.8}$  &21.6$_{3.4}$ &26.5$_{1.8}$ & 21.6$_{3.4}$  &29.0$_{1.0}$ &33.6$_{0.5}$ & 29.0$_{1.0}$  &33.0$_{1.2}$ &37.4$_{2.2}$ & 33.0$_{1.2}$ \\
CoDL        &7.7$_{1.2}$ &8.9$_{1.9}$ &4.8$_{1.1}$    &12.8$_{3.0}$ &14.2$_{3.5}$ &12.3$_{6.2}$  &16.2$_{3.1}$ &17.7$_{3.3}$ &17.3$_{2.7}$   &17.6$_{1.4}$ &18.4$_{1.8}$ &17.7$_{2.1}$   &27.0$_{3.7}$ &28.3$_{3.9}$ &22.4$_{5.0}$   &32.9$_{1.7}$ &34.6$_{0.7}$ &32.4$_{ 2.}$9  &37.2$_{1.4}$ &38.5$_{1.7}$ &35.3$_{3.7}$\\
t-norm      &8.9$_{5.1}$ &9.5$_{5.5}$ &9.5$_{3.2}$  &14.5$_{2.1}$ &15.5$_{2.4}$ &10.2$_{2.1}$  &19.2$_{5.8}$ &14.2$_{6.2}$ &19.6$_{3.1}$   &21.8$_{7.7}$ &23.1$_{7.6}$ &24.9$_{0.6}$   &\B30.1$_{1.0}$ &30.7$_{1.1}$ &29.4$_{3.9}$   &34.1$_{2.7}$ &35.6$_{3.3}$ &37.3$_{0.7}$   &37.4$_{2.5}$ &38.6$_{3.0}$ &39.5$_{1.4}$\\
sl (ours)   &\B12.9$_{3.1}$ &\B14.0$_{3.4}$ &\B14.1\rlap{$^*$}$_{3.5}$ &\B20.3$_{2.2}$ &\B22.3\rlap{$^*$}$_{3.7}$ &\B18.5$_{1.5}$  &\B24.6$_{4.1}$ &\B25.6$_{4.5}$ &\B26.2\rlap{$^*$}$_{2.1}$   &\B30.1$_{1.9}$ &\B31.3$_{2.7}$ &\B31.6\rlap{$^*$}$_{0.7}$   &29.2$_{1.7}$ &\B31.9$_{0.9}$ &\B33.6\rlap{$^*$}$_{3.3}$   &\B36.5$_{1.4}$ &\B39.4$_{1.2}$ &\B40.5\rlap{$^*$}$_{1.3}$   &\B38.8$_{1.4}$ &\B42.0\rlap{$^*$}$_{0.5}$ &\B40.3$_{0.7}$\\
\bottomrule
\end{tabular}%
}
\vskip -0.1in
\caption{Experimental results for joint entity-relation extraction on ACE05. The upper and lower sub-tables correspond to \textit{avg-f1} and \textit{tri-f1}, respectively. \textbf{Inf} indicates no constraints during inference (N), use of ILP during inference (I) or the transductive setting (T). We averages across 3 different runs. Best results in a column and block are boldface and asterisked, respectively.}
\end{table*}
\section{SciERC Experiments}
\paragraph{Dataset and Evaluation Criteria}
We conduct further experiments using the SciERC dataset \cite{luan2018}, which includes annotations for scientific entities and there relations, assimilated from 12 AI conference/workshop proceedings in four AI communities from the Semantic Scholar Corpus. The dataset defines six entity types with 7 possible relation between them. 

\paragraph{Constraint} Unlike ACE05, SciERC does not specify an ontology of entities and their permissible relations. Therefore, our constraint is determined through procuring the set of all possible relation-subject-object triples in the training set, and applying a threshold to eliminate all noisy labelings in the training set.
\begin{table*}[!htbp]
\centering
\resizebox{\columnwidth}{!}{%
\begin{tabular}{lccc|ccc|ccc|ccc|ccc|ccc|ccc}
\toprule
\# Labels      &\multicolumn{3}{c}{3}  &\multicolumn{3}{c}{5}  &\multicolumn{3}{c}{10} &\multicolumn{3}{c}{15} &\multicolumn{3}{c}{25} &\multicolumn{3}{c}{50} &\multicolumn{3}{c}{75}\\
{Inf}     &N &I &T                &N &I &T                &N &I &T                &N &I &T                &N &I &T                &N &I &T                &N &I &T\\
\midrule
base
&25.7$_{3.7}$ &46.1$_{2.6}$ &25.7$_{3.7}$
&30.6$_{7.4}$ &50.3$_{6.6}$ &30.6$_{7.4}$
&34.9$_{2.0}$ &56.6$_{2.3}$ &34.9$_{2.0}$
&36.9$_{3.1}$ &58.5$_{5.1}$ &36.9$_{3.1}$
&43.2$_{0.9}$ &61.6$_{2.6}$ &43.2$_{0.9}$
&48.9$_{2.0}$ &65.3$_{2.4}$ &48.9$_{2.0}$
&50.8$_{3.2}$ &68.2$_{1.5}$ &50.8$_{3.2}$\\
CoDL
&35.1$_{7.9}$ &53.1$_{9.0}$ &40.2$_{13.}$1
&31.0$_{7.2}$ &50.9$_{6.3}$ &29.7$_{8.0}$
&39.5$_{4.2}$ &60.2$_{3.6}$ &37.9$_{1.3}$
&37.1$_{3.4}$ &58.7$_{5.2}$ &37.9$_{2.1}$
&48.1$_{4.7}$ &66.3$_{2.0}$ &47.8$_{1.1}$
&52.9$_{0.7}$ &68.2$_{1.3}$ &51.7$_{1.2}$
&54.4$_{4.4}$ &70.9$_{2.0}$ &53.1$_{4.9}$\\  
t-norm
&42.3$_{3.8}$ &59.8$_{1.9}$ &\B43.3$_{1.3}$
&46.3$_{2.0}$ &\B62.3\rlap{$^*$}$_{0.7}$ &46.3$_{2.2}$
&49.0$_{2.3}$ &63.7$_{0.8}$ &49.4$_{2.1}$
&51.4$_{0.7}$ &64.1$_{0.8}$ &\B51.1$_{1.4}$
&52.5$_{2.8}$ &66.7$_{2.1}$ &\B53.8$_{0.8}$
&58.0$_{1.7}$ &71.5$_{1.1}$ &59.0$_{0.9}$
&57.6$_{0.9}$ &71.7$_{0.7}$ &58.0$_{1.6}$\\  
sl (ours)
&\B43.0$_{3.1}$ &\B60.6\rlap{$^*$}$_{0.4}$ &41.9$_{0.5}$
&\B47.2$_{2.4}$ &61.9$_{0.1}$ &\B47.7$_{2.1}$
&\B49.3$_{1.5}$ &\B63.9\rlap{$^*$}$_{0.5}$ &\B50.0$_{2.0}$
&\B51.8$_{1.6}$ &\B65.2\rlap{$^*$}$_{1.5}$ &49.3$_{5.1}$
&\B54.2$_{1.6}$ &\B66.8\rlap{$^*$}$_{1.6}$ &53.7$_{4.4}$
&\B62.2$_{1.2}$ &\B74.0\rlap{$^*$}$_{0.4}$ &\B60.2$_{0.6}$
&\B60.2$_{2.3}$ &\B73.2\rlap{$^*$}$_{2.0}$ &\B61.2$_{2.5}$\\
\midrule
\midrule
base
&2.7$_{1.1}$ &3.1$_{0.9}$ &2.7$_{1.1}$
&2.9$_{1.0}$ &3.2$_{0.8}$ &2.9$_{1.0}$
&3.5$_{1.8}$ &3.9$_{1.9}$ &3.5$_{1.8}$
&3.6$_{1.1}$ &4.1$_{1.8}$ &3.6$_{1.1}$
&8.8$_{1.0}$ &9.8$_{1.4}$ &8.8$_{1.0}$
&12.3$_{3.0}$ &13.4$_{3.3}$ &12.3$_{3.0}$
&12.5$_{2.6}$ &13.7$_{2.9}$ &12.5$_{2.6}$\\
CoDL
&3.6$_{1.4}$ &3.7$_{1.3}$ &\B7.9\rlap{$^*$}$_{5.9}$
&3.0$_{0.9}$ &3.4$_{0.7}$ &2.6$_{1.3}$
&4.1$_{2.6}$ &4.3$_{2.6}$ &4.6$_{2.2}$
&3.7$_{1.1}$ &4.3$_{1.8}$ &3.9$_{1.6}$
&9.4$_{3.8}$ &9.6$_{3.9}$ &10.1$_{2.0}$
&14.8$_{1.2}$ &15.5$_{1.5}$ &14.0$_{2.0}$
&13.8$_{3.9}$ &14.2$_{4.1}$ &13.1$_{5.2}$\\
t-norm
&6.5$_{2.0}$ &6.6$_{2.0}$ &6.2$_{1.3}$
&8.9$_{1.2}$ &9.0$_{1.3}$ &9.1$_{2.3}$
&10.9$_{1.6}$ &11.2$_{1.6}$ &11.3$_{2.0}$
&13.4$_{0.7}$ &13.8$_{1.0}$ &\B13.9$_{1.2}$
&13.8$_{2.9}$ &14.3$_{2.7}$ &15.0$_{1.2}$
&19.2$_{1.7}$ &19.8$_{1.8}$ &20.1$_{1.7}$
&19.5$_{1.7}$ &20.4$_{2.0}$ &19.4$_{1.6}$\\
sl (ours)
&\B6.7$_{1.5}$ &\B6.8$_{1.6}$ &6.5$_{1.5}$
&\B10.4$_{1.9}$ &\B10.6$_{1.9}$ &\B11.1\rlap{$^*$}$_{0.9}$
&\B11.1$_{1.9}$ &\B11.5$_{1.9}$ &\B12.6\rlap{$^*$}$_{2.0}$
&\B14.1$_{1.0}$ &\B14.7\rlap{$^*$}$_{0.9}$ &13.8$_{3.6}$
&\B14.4$_{1.8}$ &\B14.8$_{2.1}$ &\B17.0\rlap{$^*$}$_{3.3}$
&\B23.2$_{2.3}$ &\B23.5\rlap{$^*$}$_{2.4}$ &\B21.5$_{1.8}$
&\B20.8$_{2.3}$ &\B21.6$_{2.3}$ &\B23.8\rlap{$^*$}$_{0.7}$\\
\bottomrule
\end{tabular}%
}
\vskip -0.1in
\caption{Experimental results for joint entity-relation extraction on sciERC. The upper and lower sub-tables correspond to \textit{avg-f1} and \textit{tri-f1}, respectively. \textbf{Inf} indicates no constraints during inference (N), use of ILP during inference (I) or the transductive setting (T). We averages across 3 different runs. Best results in a column and block are boldface and asterisked, respectively.}
\end{table*}

\section{Hyperparameters}
We performed an initial grid search over optim $ \in \{\text{Adam}, \text{SGD}\}$, learning rate $\in \{10^{-3}, 10^{-2}, 10^{-1}\}$, batch size $\in \{32, 64, 128, 256\}$, semantic weight $\in \{0.001, 0.005, 0.01, 0.05, 0.1, 0.5, 1\}$ and t-norm weight $\in \{0.001, 0.005, 0.01, 0.05, 0.1, 0.5, 1, 2, 4\}$ to establish a good neighborhood of hyperparameters, which showed that while slower to converge, models trained using SGD frequently outperformed models trained using Adam, as measured by their respective performance on the validation dataset, and that a learning rate of $1.0$ was a universally good learning rate for our architecture. Therefore, for all our subsequent experiments, we fixed SGD as the optimizer, and $1.0$ as the initial learning rate, which was annealed by a decay rate of $0.9$ for every 10 epochs that the model did not make progress, as measured by its joint performance on the entity-relation extraction task. Every model is allowed to train for $100$ epochs, with early stopping if progress is not made for $20$ epochs. All experiments were conducted on a GeForce GTX 1080 Ti GPU.
\section{Implementation and Training Details}
 Our model is implemented in PyTorch \cite{Paszke19} using the BERT$_{BASE}$ model from the PyTorch Transformers library\footnote{https://github.com/huggingface/transformers}. Our code makes use of the PySDD\footnote{https://github.com/wannesm/PySDD} as well as the PyPSDD\footnote{https://github.com/art-ai/pypsdd} libraries for compiling logical constraints into circuits which enable the efficient calculation of semantic loss.
 
 \paragraph{Unlabeled Data} Both semantic loss and product t-norm are provided with the entire, unlabeled corpus. where as, inline with previous work \citep{chang2007}, CoDL is provided as input the portion of the corpus not contained within the labeled set.
 
 \paragraph{Constraint} The ACE05 specification lists all permissible relations and their arguments \footnote{https://www.ldc.upenn.edu/sites/www.ldc.upenn.edu/files/english-relations-guidelines-v5.8.3.pdf, section 3}, the domain knowledge, which we refer to as our ontology or constraint through out our paper.
 
 \end{document}